%% file: main.tex
\documentclass[10pt,twocolumn,letterpaper]{article}

\usepackage{cvpr}              
\input{preamble}
\definecolor{cvprblue}{rgb}{0.21,0.49,0.74}
\usepackage[pagebackref,breaklinks,colorlinks,allcolors=cvprblue]{hyperref}

\def\confName{CVPR}
\def\confYear{2026}

\title{Reinforcing Egocentric Spatial Perception in Multimodal Large Language Models via Ego Scene Augmentation}

\author{%
Chi Kit Wong$^{1,*}$ \quad Ye Pan$^{1,*}$ \quad Yuanhuiyi Lyu$^{1}$\\
Xu Zheng$^{1}$ \quad Zidong Cao$^{1}$ \quad Lutao Jiang$^{1}$\\
Zixin Zhang$^{1}$ \quad Huiyu Zhou$^{2}$ \quad Xuming Hu$^{1,3}$\\[1mm]
$^{1}$The Hong Kong University of Science and Technology (Guangzhou)\\
$^{2}$Guangxi Zhuang Autonomous Region Information Center\\
$^{3}$The Hong Kong University of Science and Technology\\[0.2mm]
Code Repository: \href{https://github.com/Chikit-WONG/spatialGraph}{\tt\small https://github.com/Chikit-WONG/spatialGraph}\\[0.2mm]
{\scriptsize\texttt{\{ckwong627,ypan477\}@connect.hkust-gz.edu.cn}, \quad
\texttt{ryan.lyu.mail@gmail.com}, \quad
\texttt{xuminghu@hkust-gz.edu.cn}}
}

\usepackage{tikz}
\usetikzlibrary{positioning,arrows.meta,calc}

\usepackage{booktabs}   
\usepackage{multirow}   
\usepackage{colortbl}   
\usepackage{xcolor}     
\usepackage{graphicx}   
\usepackage{xcolor}
\usepackage{amsmath}

\usepackage{float}      

\usepackage{caption}
\usepackage{subcaption} 

\usepackage{algorithm}

\usepackage{algorithm}
\usepackage[noend]{algpseudocode} 
\usepackage{amsmath}

\newcommand{\aka}{\textit{a.k.a.}}

\begin{document}

\maketitle
\begingroup
\renewcommand{\thefootnote}{\fnsymbol{footnote}}
\footnotetext[1]{Equal contribution.}
\endgroup
\input{sec/0_abstract}    
\input{sec/1_intro}

\input{sec/2_related-work}

\input{sec/3_methodology}
\input{sec/4_experiment}
\input{sec/5_ablation_study}

\input{sec/6_conclusion}

\clearpage
\newpage
{
    \footnotesize
    \bibliographystyle{ieeenat_fullname}
    \bibliography{main}
}


\input{sec/X_suppl}

\end{document}

%% file: sec/0_abstract.tex
\begin{abstract}

Egocentric Visual Question Answering (VQA) has attracted widespread attention as an important task for enabling Multimodal Large Language Models (MLLMs) to interact with the real world.
However, existing MLLMs struggle to perform effective spatial reasoning in complex egocentric scenes due to their limited spatial perception capabilities.
To this end, we introduce \textbf{Ego Scene Augmentation (ESA)}, an egocentric spatial perception framework, which actively enhances the spatial perception capabilities from the egocentric perspective, powered by the proposed Ego-element Graph.
Our \textbf{core insight} is leveraging the Ego-element Graph as an intermediary representation to augment the egocentric spatial perception of MLLMs via visual foundational models.
Specifically, we \textbf{1)} construct the Ego-element Graph, which encapsulates and integrates egocentric spatial features enabled by visual foundational models; \textbf{2)} enhance the spatial perception capabilities of MLLMs via the Ego-element Graph for ego-perspective scenes.
Our proposed ESA framework presents significant performance improvement on the EgoTextVQA~\cite{zhou2025egotextvqa} benchmark. We achieve an \textbf{8.14\%} gain on the indoor setting and an \textbf{8.72\%} gain on the outdoor setting. Furthermore, our ESA shows the most impressive performance improvement in the shopping subset of the indoor setting.
\textit{The project code is publicly available.}

\end{abstract}

\begin{figure}
    \centering
    \includegraphics[width=1\linewidth]{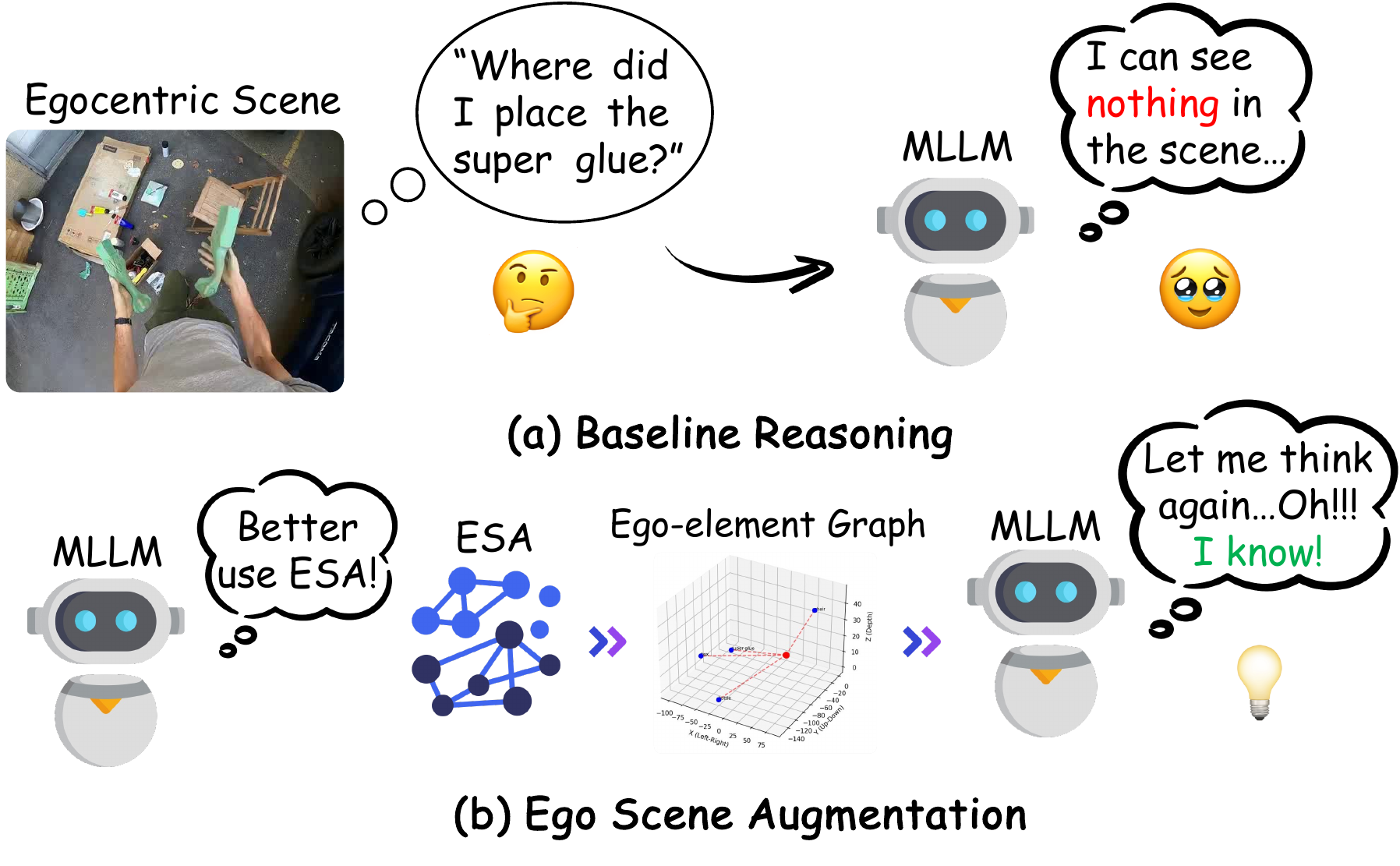}
    \caption{(a) In spatially complex egocentric scenes, current MLLMs lack spatial perception capabilities, often mislocalize objects, and therefore produce an “Unanswerable” output when ESA is absent.
(b) With ESA, Ego-element Graph is incorporated into inference, supporting reliable localization and leading to more accurate, well grounded answers.}
    \label{fig:coverage}
\end{figure}

%% file: sec/1_intro.tex
\section{Introduction}
\label{sec:intro}


Recent multimodal models~\cite{wang2024emu3,chen2024internvl,bai2025qwen2,wang2024qwen2,dai2023instructblip,li2024llava} have demonstrated remarkable performance in visual understanding. However, MLLMs remain limited in the context of egocentric VQA, where they rely heavily on pre-trained visual perception capabilities.
Specifically, \textbf{(1)} these models struggle to adapt their perception mechanisms to the egocentric viewpoint required for egocentric VQA; \textbf{(2)} it is challenging for them to conduct accurate spatial perception in the highly dynamic and cluttered egocentric scenes.



In this paper, we propose the \textit{Ego Scene Augmentation} (\textbf{ESA}), a novel egocentric spatial perception framework that explicitly enhances the spatial perception capabilities of MLLMs from an egocentric perspective. ESA improves egocentric spatial perception through a proposed Ego-element Graph, constructed using visual foundation models.
The core idea is to utilize the Ego-element Graph as an intermediate representation to augment the egocentric spatial perception of MLLMs via integration with visual foundation models.
In this way, our ESA effectively infuses the spatial abilities of visual foundation models into the MLLM and further enhances their performance in the egocentric scene with a ``\textit{plug-and-play}" approach.


To this end, we first construct the Ego-element Graph, powered by a visual foundation model—specifically, \aka, Depth-Anything~\cite{yang2024depth}. We then employ this graph as an intermediate representation that restores and transfers the visual perception capabilities of the visual foundation model to MLLMs, thereby enhancing their egocentric visual understanding.
Notably, ESA is designed with a ``\textit{plug-and-play}" philosophy, enabling flexible integration into existing MLLMs. In this way, ESA effectively augments egocentric spatial perception in a modular and adaptable manner.
As shown in Fig.~\ref{fig:coverage}, compared to the original MLLMs, our Ego-element Graph significantly improves their ability to identify key regions and spatial relationships within egocentric scenes, and further guides them toward more accurate responses for egocentric VQA.

Based on the ``\textit{plug-and-play}" design, our ESA can be applied with existing MLLM flexibly, \eg, Emu3~\cite{wang2024emu3}, LLaVA-NeXT~\cite{li2024llava}, InternVL2~\cite{chen2024internvl}, Qwen2-VL~\cite{wang2024qwen2}, Qwen2.5-VL~\cite{bai2025qwen2}.
Our ESA also presents significant performance gains on the EgoTextVQA~\cite{zhou2025egotextvqa} benchmark via the Ego-element Graph. Specifically, we achieve an \textbf{8.72\%} performance improvement on the outdoor setting, and an \textbf{8.14\%} performance improvement on the indoor setting.

\noindent \textit{Our main contributions are summarized as follows:}
\begin{itemize}
    \item We propose ESA, a novel egocentric spatial perception framework that constructs an Ego-element Graph using visual foundation models. This graph serves as an explicit egocentric spatial representation to enhance the spatial perception capabilities of MLLMs.
    \item We design ESA as a ``\textit{plug-and-play}", model-agnostic module that can be flexibly integrated into existing MLLMs. By using the Ego-element Graph as an intermediate representation, ESA effectively infuses the spatial perception of visual foundation models into MLLMs.
    \item Extensive experiments on the EgoTextVQA~\cite{zhou2025egotextvqa} benchmark demonstrate that ESA brings significant performance gains in egocentric VQA, improving accuracy by 8.72\% in outdoor scenes and 8.14\% in indoor scenes, validating the effectiveness and generality of our approach.
\end{itemize}

%% file: sec/2_related-work.tex
\section{Related Work}
\label{sec:formatting}

\subsection{Egocentric VQA}

Egocentric visual question answering (VQA) has gradually evolved from third-person visual understanding to embodied, first-person reasoning. Early works mainly focused on static image understanding and text recognition~\cite{antol2015vqa, wu2017visual, singh2019towards, biten2019scene}, which lacked the temporal and spatial grounding necessary for egocentric settings. To bridge this gap, datasets such as EgoVQA~\cite{fan2019egovqa}, EgoTaskQA~\cite{jia2022egotaskqa}, and EMQA~\cite{barmann2022did} extended VQA into short video segments, emphasizing temporal reasoning and causal understanding. However, their scenes were still relatively simple, often limited to constrained household environments.

Recent advances have brought richer and more diverse benchmarks. EgoCross~\cite{li2025egocross} and EgoLife~\cite{yang2025egolife} introduced cross-domain generalization and long-horizon reasoning across daily activities, while EgoNight~\cite{zhang2025egonight} examined robustness under extreme lighting variations. More recent datasets such as EgoThink~\cite{cheng2024egothink}, VidEgoThink~\cite{cheng2024videgothink}, and MM-Ego~\cite{ye2024mm} have further emphasized multi-modal perception and memory-based reasoning, pushing the field toward fully embodied understanding. Meanwhile, benchmarks like AlanaVLM~\cite{suglia2024alanavlm} and ESVQA~\cite{zhu2024esvqa} have begun to integrate egocentric perception into assistive and interactive systems, especially in accessibility scenarios.

Despite these developments, prior benchmarks rarely focus on text-rich egocentric environments, where understanding scene text, instructions, and object relations is crucial for reasoning. EgoTextVQA~\cite{zhou2025egotextvqa} fills this gap by introducing an egocentric, text-intensive benchmark that jointly evaluates scene-text awareness, temporal localization, key-frame reasoning, and tri-modal alignment among vision, scene text, and language. This benchmark highlights the need for models capable of spatially grounded reasoning in dynamic and cluttered first-person scenes.

Although recent multi-modal large language models~\cite{wang2024emu3, chen2024internvl, bai2025qwen2, wang2024qwen2, dai2023instructblip, li2024llava} have demonstrated impressive generalization across visual and linguistic tasks, they continue to struggle on egocentric benchmarks. The main challenges stem from unstable spatial grounding, limited temporal alignment, and insufficient integration of text–object relations. To address these limitations, we propose ESA, a plug-and-play module that explicitly constructs structured spatial–object–text graphs from egocentric scenes. This design enhances spatial reasoning, mitigates attention drift, and significantly improves model performance on text-driven egocentric benchmarks such as EgoTextVQA~\cite{zhou2025egotextvqa}.

\paragraph{Multimodal Representation and Knowledge Augmentation.}
Recent studies augment multimodal systems through improved representation spaces and external visual knowledge. UniBind~\cite{lyu2024unibind} constructs LLM-augmented class-wise embedding centers to learn a modality-agnostic, unified, and balanced representation space across diverse modalities. RealRAG~\cite{lyu2025realrag} uses a self-reflective retriever to retrieve real-world reference images that supply missing knowledge to text-to-image generative models. In contrast, ESA targets egocentric VQA and provides explicit spatial knowledge at inference time through an Ego-element Graph.

\subsection{Monocular Depth Estimation}

Monocular depth estimation (MDE)~\cite{roy2016monocular, godard2017unsupervised, fu2018deep, wofk2019fastdepth, li2022binsformer, yang2024depth, bochkovskii2024depth, piccinelli2025unidepthv2, bhat2023zoedepth, bhat2021adabins, ranftl2020towards, yin2023metric3d, bhat2022localbins, mertan2022single, gui2025depthfm, fu2024geowizard, hu2024metric3d, guizilini2023towards, li2024patchfusion, ke2024repurposing, yuan2022new, godard2019digging} aims to infer pixel-wise depth from a single RGB image, serving as a fundamental component of modern 3D perception and reliable spatial reasoning. Early approaches, such as Neural Regression Forests~\cite{roy2016monocular}, combined CNNs and regression trees to infer continuous depth, while unsupervised frameworks like Godard \textit{et al.}~\cite{godard2017unsupervised, godard2019digging} replaced dense depth supervision with stereo image reconstruction losses based on left-right consistency, significantly reducing the dependence on labeled data.

Subsequent methods reformulated depth estimation to enhance optimization stability and accuracy. The Deep Ordinal Regression Network~\cite{fu2018deep} introduced spacing-increasing discretization to model depth as an ordinal regression problem, achieving superior convergence and accuracy. To improve efficiency, FastDepth~\cite{wofk2019fastdepth} employed a lightweight MobileNet-based encoder–decoder architecture for real-time inference on embedded systems. Later, AdaBins~\cite{bhat2021adabins}, LocalBins~\cite{bhat2022localbins}, and BinsFormer~\cite{li2022binsformer} refined adaptive binning strategies, while transformer-based models such as Metric3D~\cite{yin2023metric3d, hu2024metric3d}, PatchFusion~\cite{li2024patchfusion}, and GeoWizard~\cite{fu2024geowizard} improved depth distribution learning.

More recent efforts~\cite{ranftl2020towards, guizilini2023towards, bhat2023zoedepth, ke2024repurposing, gui2025depthfm, piccinelli2025unidepthv2} emphasize cross-domain generalization and foundation-level scalability. Depth Anything V2~\cite{yang2024depth} represents a new milestone, replacing all real-labeled data with high-fidelity synthetic supervision in a comprehensive teacher–student distillation framework. By bridging the synthetic-to-real domain gap through large-scale pseudo-labeled real images, it achieves fine-grained, stable, and highly transferable depth estimation across diverse visual domains.

In our framework, Depth Anything V2~\cite{yang2024depth} serves as the backbone for spatial understanding, providing reliable geometric cues for constructing the Ego-element Graph. Its robust performance under motion blur, temporal shifts, and low-texture conditions makes it particularly suited for egocentric scenarios. By integrating accurate depth priors into the Ego-element Graph, we enhance spatial reasoning and object–text grounding in egocentric VQA tasks.

%% file: sec/3_methodology.tex
\section{Methodology}
\label{sec:methodolgy}


\subsection{Problem Setting and Overview}

\noindent \textbf{Problem Setting.}
Given a textual question $Q$ and a scene image $I$, the Egocentric VQA task aims to respond with an answer $Ans$ that accurately reflects a conditional distribution $P\left( Ans \mid Q,\,I \right)$, which describes the likelihood that an answer $Ans$ is given to the question $Q$ and the egocentric scene $I$. In practice, we often parameterize this distribution slightly more explicitly related to an MLLM with parameters $\theta$, denoted by $M_{\theta}$, yielding:

\begin{equation}
P_{\theta}\left(Ans \mid Q,\,I\right) \approx P\left(Ans \mid Q,\,I\right)\,.
\label{eq:conditional_prob_with_theta}
\end{equation}
where $P_{\theta}$ is the conditional distribution related to the $M_{\theta}$.

Given a new text question $Q$ and a scene image $I$, the well-trained MLLM $M_{\theta}$ respond with an answer:
\begin{equation}
    \hat{Ans} \sim M_{\theta}\left(Q,\,I\right)\,.
    \label{eq:sample_generation}
\end{equation}

The task goal is to ensure that $\hat{Ans}$ aligns with the visual evidence in $I$ and accurately answers the question $Q$.

\begin{figure}[htbp]
  \centering
  \includegraphics[width=\linewidth]{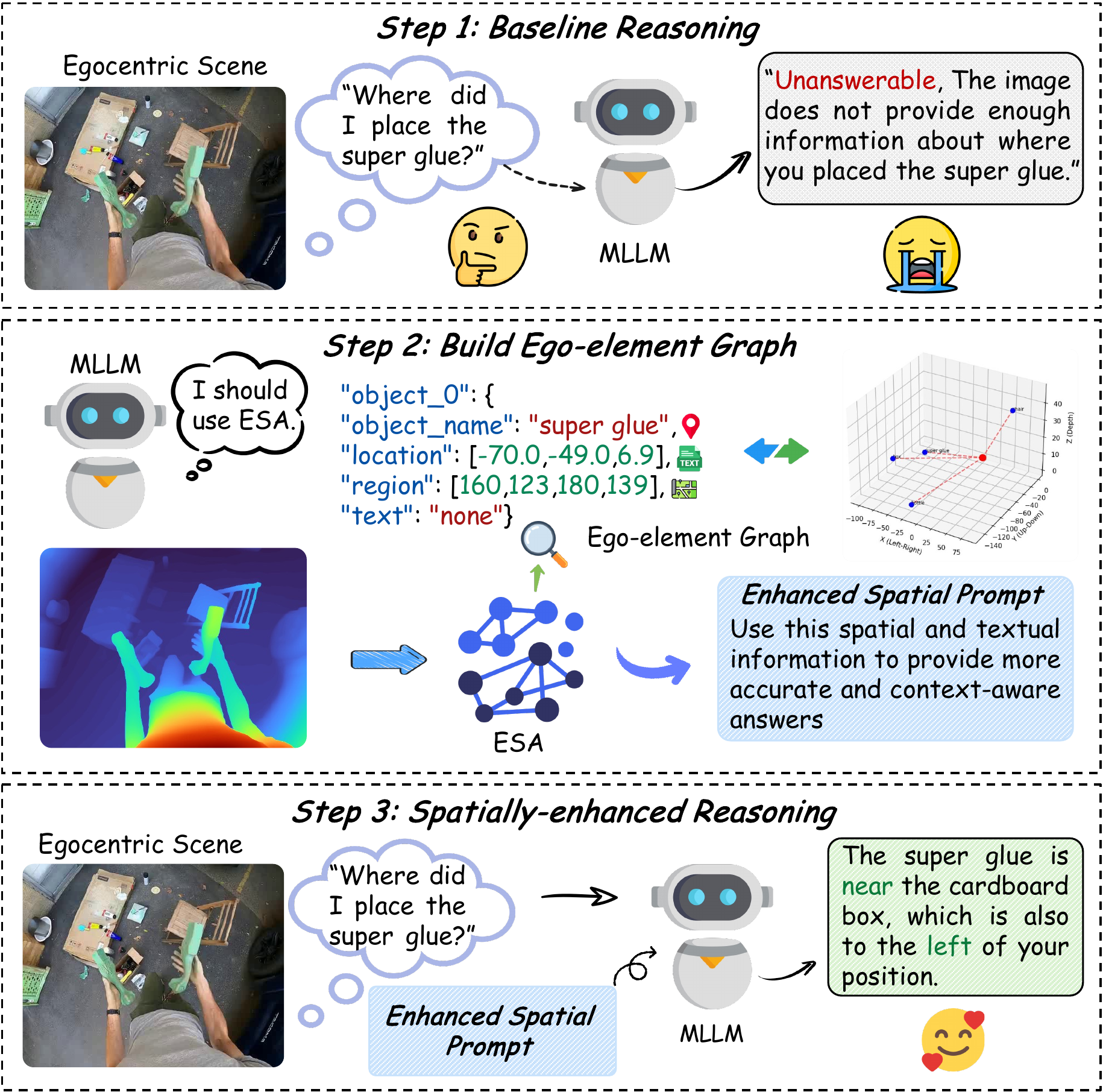}
  \caption{The overall framework of our proposed Ego Scene Augmentation for the Egocentric VQA task.}
  \label{fig:whole_pipeline}
\end{figure}

\noindent \textbf{Overview.} As illustrated in Figure~\ref{fig:whole_pipeline},  we present ESA, a spatially grounded reasoning framework that improves multi-modal large language model in egocentric scenes. To address the limitation of egocentric spatial perception in MLLM, we construct the Ego-element Graph that encodes object semantics, 3D positions, and visual regions. The resulting Ego-element Graph is provided to the MLLM as an enhanced prompt, enabling more accurate, context-aware perception. With this added Ego-element Graph, the model can infer object placement, \eg, locating the super glue relative to the user, thus significantly improving egocentric spatial understanding.
The ``\textit{key insight}" of our ESA is to utilize the Ego-element Graph as an intermediate representation to augment the egocentric spatial perception of MLLM via integration with visual foundation models.

\input{tables/egotextvqa_result_indoor}
\input{tables/egotextvqa_result_outdoor}

\subsection{Ego-element Graph}

To enable reliable spatial reasoning in egocentric scenes, we introduce the \textit{Ego-element Graph}, a structured representation that integrates object-level semantics, geometry, and visual context. Formally, an Ego-element Graph is defined as a directed attributed graph:
\begin{equation}
    G = (V, E, A),
\end{equation}
where $V$ is the set of nodes, each representing an object or salient region in the egocentric scene; $E$ is the set of directed edges encoding pairwise spatial and contextual relations between ego-elements; and $A$ denotes the set of attribute vectors associated with all nodes and edges.

\noindent \textbf{Nodes.} Each node $v_i \in V$ corresponds to an \textit{ego-element}, representing an object or salient region in the scene. An ego-element is characterized by a unified attribute vector:
\begin{equation}
    a_i = \left[\, s_i \;\Vert\; p_i \;\Vert\; r_i \;\Vert\; t_i \,\right],
\end{equation}
where $s_i$ denotes semantic features (\eg, element names), $p_i \in \mathbb{R}^3$ is the 3D position of the object in the egocentric scene, $r_i \in \mathbb{R}^4$ represents the 2D region, $t_i$ encodes associated textual cues.
Thus, each node is defined as:
\begin{equation}
    v_i = (i, a_i).
\end{equation}

\noindent \textbf{Edges.} Each edge $e_{ij} \in E$ encodes \textit{pairwise spatial and contextual relations} between ego-elements. For any pair $(v_i, v_j)$, we compute a relation vector:
\begin{equation}
    {r}_{ij} = \left[\, {p}_j - {p}_i \;\Vert\; \phi({r}_i, {r}_j) \,\right],
\end{equation}
where ${p}_j - {p}_i$ represents relative depth displacement and
$\phi({r}_i, {r}_j)$ encodes region-based relations. 
To further adapt to complex egocentric scenes, our edges are directed to reflect user-centric orientation, \eg, ``\textit{left of}", ``\textit{in front of}".

Given an egocentric observation $I$, the proposed ESA $f_{esa} ( \cdot ) $ extracts all ego-elements and constructs:
\begin{equation}
    G = f_{\text{esa}}(I),
\end{equation}
yielding the egocentric scene-level Ego-element Graph that fuses geometric, visual, and textual evidence.

\noindent \textbf{Enhanced Spatial Prompt.} Finally, the graph is serialized into a prompt representation:
\begin{equation}
    Prompt_{spatial} = Encoder(G),
\end{equation}
where $Encoder$ means a JSON structure encoder to transfer the Ego-element Graph into the JSON text prompt.
The enhanced prompt is provided to an MLLM to guide spatially aware perception for egocentric VQA.

\begin{figure*}[t!]
  \centering
  \includegraphics[width=\textwidth]{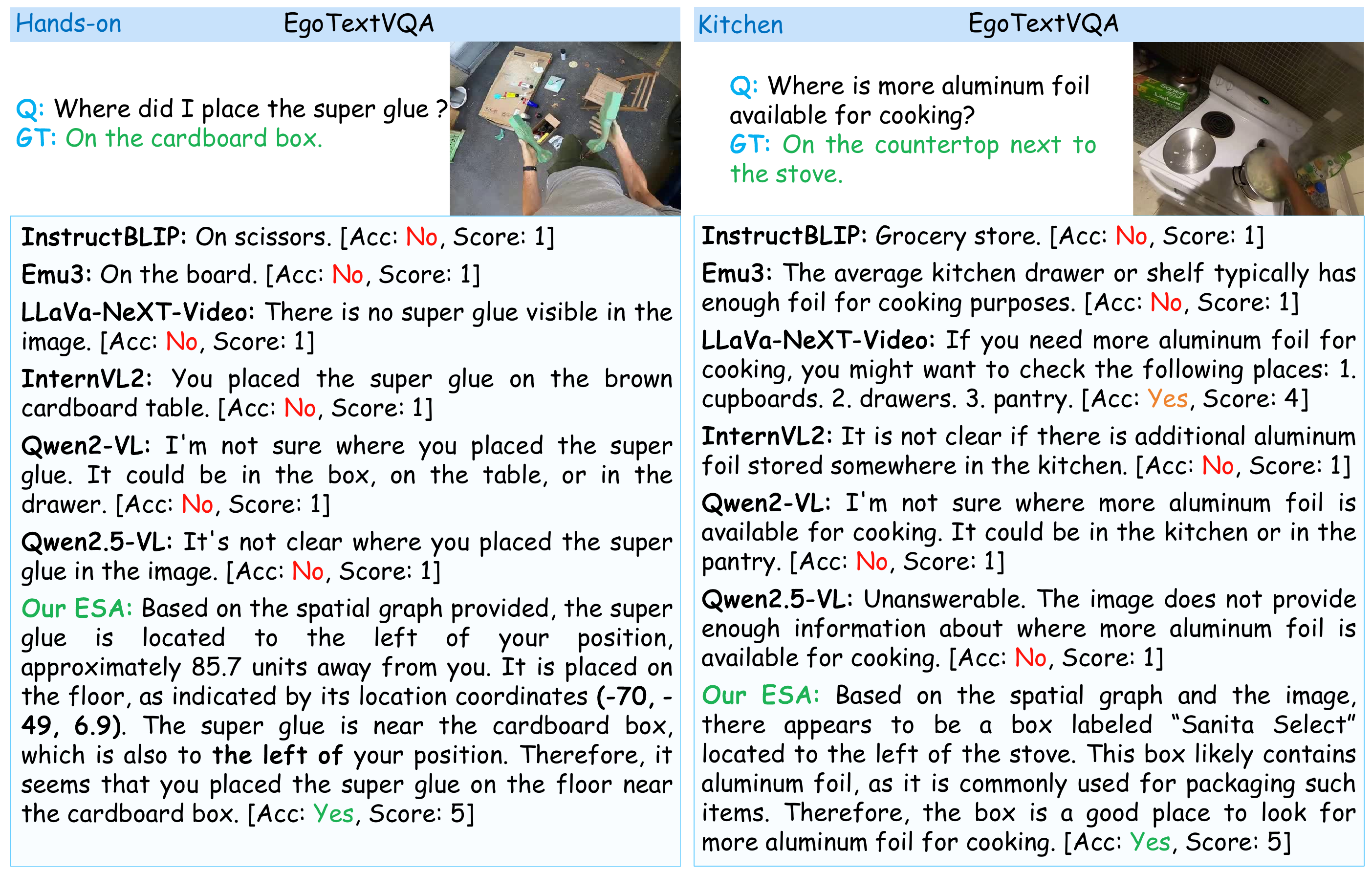}
  \vspace{-12pt}
  \caption{Comparison of existing MLLMs and our Ego Scene Augmentation (ESA) on EgoTextVQA.
ESA leverages the constructed Ego-element Graph to answer queries that baseline MLLMs fail to handle, producing more accurate and spatially grounded predictions across diverse scenarios (e.g., Hands-on and Kitchen).}
  \label{fig:experiment qualitative results}
\end{figure*}

\subsection{``Plug-and-play" Framework}






Our ESA framework is designed as a plug-and-play module that augments any existing MLLMs without any training.
Given a question $Q$ and the egocentric scene $I$, our ESA first builds the proposed Ego-element Graph $G$ and then serializes it into an enhanced spatial prompt to effectively and robustly augment the egocentric spatial perception by:
\begin{equation}
    \hat{Ans} = MLLM(Q, Prompt_{spatial}).
\end{equation}

Based on the ``\textit{plug-and-play}" design, our ESA can be applied with existing MLLMs flexibly, \eg Qwen2.5-VL~\cite{bai2025qwen2}.
Our ESA also presents significant performance gains on the EgoTextVQA~\cite{zhou2025egotextvqa} benchmark via the Ego-element Graph. Specifically, we achieve an \textbf{8.72\%} performance improvement on the outdoor setting, and an \textbf{8.14\%} performance improvement on the indoor setting.



\subsection{Implementation}

\noindent \textbf{Base Model.}
We adopt Qwen2.5-VL~\citep{bai2025qwen2} as our base vision-language model to perform the egocentric VQA task.
Qwen2.5-VL~\citep{bai2025qwen2} is designed to advance fine-grained perception and spatial reasoning.
Compared with prior open-source models, Qwen2.5-VL~\citep{bai2025qwen2} demonstrates superior object localization, document parsing, and video grounding capabilities, while maintaining strong linguistic performance.
In our work, this robust architecture serves as the foundation for constructing Ego-element Graph and reasoning over 3D egocentric scenes.


\noindent \textbf{Depth Estimation.}
For monocular depth estimation, we employ Depth Anything V2~\citep{yang2024depth} as the depth estimation backbone to extract dense depth maps from single key frames.
Depth Anything V2~\citep{yang2024depth} is a powerful discriminative model that achieves fine-grained and robust depth prediction by leveraging synthetic-to-real transfer learning.
In our implementation, we adopt the ViT-Base variant of Depth Anything V2~\citep{yang2024depth} to obtain pixel-level depth estimation for constructing the Ego-element Graph.


\noindent \textbf{Reasoning Prompt.}
We design a reasoning prompt to strengthen the model’s spatial reasoning and contextual understanding in complex egocentric scenes.
The base prompt instructs the model to answer from a first-person perspective, while the enhanced prompt integrates spatial-graph information to support more reliable evidence-based reasoning when visual cues are insufficient.
The full text of both prompts is provided in the \textit{Supplementary Material}.


\noindent \textbf{Maximum Number of Nodes.}
We set the maximum number of detected nodes and performed an ablation study on the hyperparameter in Table~\ref{tab:ablation_num}. Based on the experimental results, we confirm that the optimal value for the maximum number of detected nodes is 4.

%% file: tables/egotextvqa_result_indoor.tex
\newcommand{\upx}[1]{\textcolor{violet!85}{$\uparrow$#1}}
\newcommand{\downx}[1]{\textcolor{red!20!green!20!blue!50}{$\downarrow$#1}}
\newcommand{\baseval}[1]{\textcolor{gray!85}{#1}}

\begin{table*}[t]
\centering
\Large 
\caption{\textbf{Evaluation Results on the EgoTextVQA Benchmark} of baselines and our ESA on the indoor setting. ``ACC" refers to Accuracy.
To facilitate a direct comparison of performance, we set the performance of the SOTA (Qwen2.5-VL) model as \baseval{-0.00}. 
Gains relative to the SOTA model are indicated in \textcolor{violet!85}{violet}, while reductions are highlighted using \textcolor{red!20!green!20!blue!50}{blue}.}
\resizebox{\textwidth}{!}{ 
\begin{tabular}{l|cccc|cc|cc|cc|cc|cc|cc}
\midrule
\multicolumn{17}{c}{\Large\textbf{EgoTextVQA-Indoor}} \\ 
\midrule
\multirow{2}{*}{\textbf{Method}} &
\multicolumn{4}{c|}{\textbf{Average}} &
\multicolumn{2}{c|}{\textbf{Hands-on}} &
\multicolumn{2}{c|}{\textbf{Kitchen}} &
\multicolumn{2}{c|}{\textbf{Shopping}} &
\multicolumn{2}{c|}{\textbf{Gameplay}} &
\multicolumn{2}{c|}{\textbf{Book-Related}} &
\multicolumn{2}{c}{\textbf{Others}} \\ 
\cmidrule{2-17}
 & \textbf{ACC} & \textbf{$\Delta$ACC} & \textbf{Score} & \textbf{$\Delta$Score} 
 & \textbf{ACC} & \textbf{Score} & \textbf{ACC} & \textbf{Score} & \textbf{ACC} & \textbf{Score}
 & \textbf{ACC} & \textbf{Score} & \textbf{ACC} & \textbf{Score} & \textbf{ACC} & \textbf{Score} \\ 
\midrule
InstructBLIP & 28.09 & \downx{3.11} & 2.12 & \downx{0.12} & 26.26 & 2.05 & 25.45 & 2.01 & 26.54 & 2.04 & 37.54 & 2.49 & 30.96 & 2.24 & 22.89 & 1.92 \\
Emu3 & 9.16 & \downx{22.04} & 1.37 & \downx{0.87} & 8.88 & 1.35 & 12.24 & 1.49 & 6.98 & 1.29 & 9.30 & 1.37 & 10.52 & 1.42 & 6.47 & 1.26 \\
LLaVA-NeXT & 28.35 & \downx{2.85} & 2.13 & \downx{0.11} & 27.98 & 2.11 & 24.48 & 1.97 & 26.82 & 2.06 & 36.54 & 2.46 & 30.34 & 2.20 & 23.38 & 1.94 \\
InternVL2 & 28.52 & \downx{2.68} & 2.12 & \downx{0.12} & 27.65 & 2.09 & 23.88 & 1.93 & 23.74 & 1.95 & 38.21 & 2.51 & 32.51 & 2.29 & 26.87 & 2.03 \\
Qwen2-VL & 28.97 & \downx{2.23} & 2.15 & \downx{0.09} & 28.51 & 2.12 & 23.28 & 1.91 & 23.46 & 1.94 & 39.87 & 2.58 & 33.44 & 2.33 & 26.37 & 2.04 \\
Qwen2.5-VL & 31.20 & \baseval{-0.00} & 2.24 & \baseval{-0.00} & 31.81 & 2.27 & 26.87 & 2.07 & 24.58 & 1.97 & 41.53 & 2.65 & 36.02 & \textbf{2.46} & 24.88 & 1.98 \\
\rowcolor{gray!15}
Ours & \textbf{33.74} & \upx{2.54} & \textbf{2.34} & \upx{0.10} & \textbf{33.86} & \textbf{2.35} & \textbf{28.96} & \textbf{2.15} & \textbf{28.49} & \textbf{2.12} & \textbf{45.00} & \textbf{2.78} & \textbf{36.84} & 2.45 & \textbf{28.86} & \textbf{2.14} \\
\midrule
\end{tabular}
}
\label{tab:egotextvqa_result_indoor}
\end{table*}

%% file: tables/egotextvqa_result_outdoor.tex

\begin{table*}[t]
\centering
\Large
\caption{\textbf{Evaluation Results on the EgoTextVQA Benchmark} of baselines and our ESA on the outdoor setting. ``ACC" refers to Accuracy and ``Int. Reasoning" to Intention Reasoning.
To facilitate a direct comparison of performance, we set the performance of the SOTA (Qwen2.5-VL) model as \baseval{-0.00}. 
Gains relative to the SOTA model are indicated in \textcolor{violet!85}{violet}, while reductions are highlighted using \textcolor{red!20!green!20!blue!50}{blue}.}
\resizebox{\textwidth}{!}{
\begin{tabular}{l|cccc|cc|cc|cc|cc|cc}
\midrule
\multicolumn{15}{c}{\Large\textbf{EgoTextVQA-Outdoor}} \\ 
\midrule
\multirow{2}{*}{\textbf{Method}} &
\multicolumn{4}{c|}{\textbf{Average}} &
\multicolumn{2}{c|}{\textbf{Location}} &
\multicolumn{2}{c|}{\textbf{Direction}} &
\multicolumn{2}{c|}{\textbf{Description}} &
\multicolumn{2}{c|}{\textbf{Int. Reasoning}} &
\multicolumn{2}{c}{\textbf{Others}} \\ 
\cmidrule{2-15}
 & \textbf{ACC} & \textbf{$\Delta$ACC} & \textbf{Score} & \textbf{$\Delta$Score} 
 & \textbf{ACC} & \textbf{Score} & \textbf{ACC} & \textbf{Score} & \textbf{ACC} & \textbf{Score}
 & \textbf{ACC} & \textbf{Score} & \textbf{ACC} & \textbf{Score} \\ 
\midrule
InstructBLIP & 21.07 & \downx{4.16} & 1.84 & \downx{0.16} & 14.29 & 1.56 & 30.25 & 2.22 & 20.99 & 1.83 & 22.14 & 1.91 & 24.74 & 1.99 \\
Emu3 & 5.71 & \downx{19.52} & 1.24 & \downx{0.76} & 5.51 & 1.22 & 10.08 & 1.43 & 3.46 & 1.14 & 5.29 & 1.22 & 3.66 & 1.13 \\
LLaVA-NeXT & 21.00 & \downx{4.23} & 1.84 & \downx{0.16} & 13.36 & 1.53 & 30.99 & 2.26 & 21.12 & 1.84 & 22.02 & 1.90 & 26.32 & 2.05 \\
InternVL2 & 23.97 & \downx{1.26} & 1.95 & \downx{0.05} & 18.54 & 1.73 & 31.93 & 2.28 & 26.10 & 2.03 & 18.94 & 1.78 & \textbf{32.98} & \textbf{2.29} \\
Qwen2-VL & 24.92 & \downx{0.31} & 1.99 & \downx{0.01} & 20.00 & 1.79 & 32.67 & 2.32 & 26.24 & 2.04 & 21.03 & 1.85 & 31.94 & 2.23 \\
Qwen2.5-VL & 25.23 & \baseval{-0.00} & 2.00 & \baseval{-0.00} & 18.60 & 1.73 & 34.98 & \textbf{2.42} & 27.33 & 2.07 & 21.53 & 1.87 & 29.32 & 2.17 \\
\rowcolor{gray!15}
Ours & \textbf{27.43} & \upx{2.20} & \textbf{2.09} & \upx{0.09} & \textbf{20.86} & \textbf{1.82} & \textbf{37.71} & 2.25 & \textbf{28.41} & \textbf{2.12} & \textbf{24.97} & \textbf{2.02} & 31.41 & 2.24 \\
\midrule
\end{tabular}
}
\label{tab:egotextvqa_result_outdoor}
\end{table*}

%% file: sec/4_experiment.tex
\section{Experiment}
\label{sec:experiment}

\input{tables/ablation_num}
\begin{figure*}[t!]
  \centering
  \includegraphics[width=\textwidth]{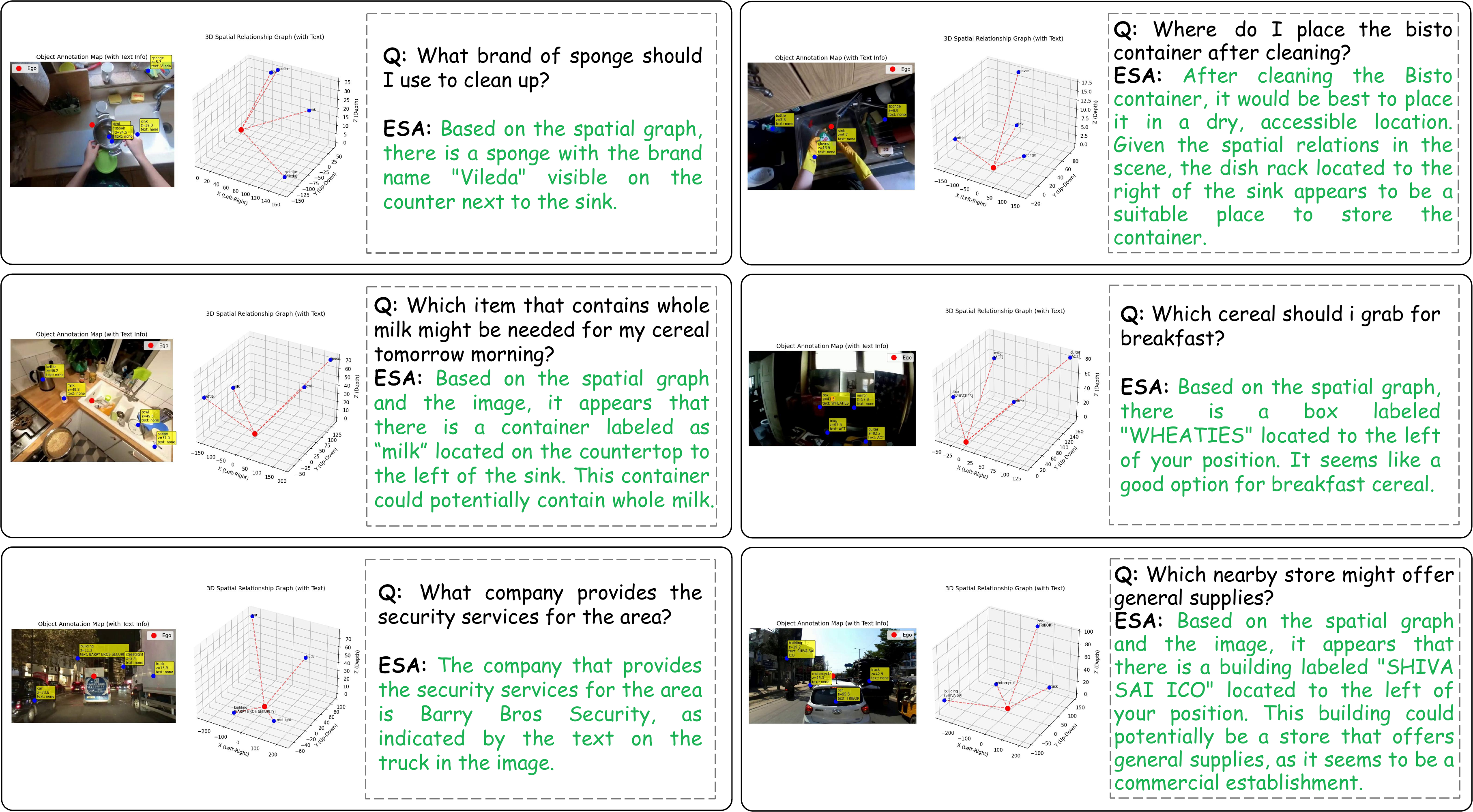}
  \caption{\textbf{Ablation on Components of the Pipeline.} Examples of model outputs after integrating the Ego-element Graph, showing how spatial and textual cues improve the reasoning process.}
  \label{fig:ablation1}
\end{figure*}

\subsection{Benchmark and Experiment Setup}

\noindent \textbf{Benchmark:}
We evaluate the performance of the egocentric VQA task using the EgoTextVQA~\cite{zhou2025egotextvqa} benchmark. It provides a comprehensive framework to evaluate the models for the egocentric VQA task which is embedded in realistic first-person visual scenes, reflecting real-world assistive scenarios. The benchmark comprises two subsets: EgoTextVQA-Indoor, covering household and object manipulation scenes from first-person daily-life recordings, and EgoTextVQA-Outdoor, featuring driving and navigation activities captured from dashcam videos. In total, the well-rounded dataset contains over 1.5K egocentric video clips and 7K question–answer pairs.

\noindent \textbf{Baselines:}
Our baseline models consist of open-source models, including InstructBLIP~\cite{dai2023instructblip}, Emu3~\cite{wang2024emu3}, LLaVA-NeXT~\cite{li2024llava}, InternVL2~\cite{chen2024internvl}, Qwen2-VL~\cite{wang2024qwen2}, and Qwen2.5-VL~\cite{bai2025qwen2}, representing diverse modern architectures.

\noindent \textbf{Experiment Setting:}
We follow the official setting of the EgoTextVQA~\cite{zhou2025egotextvqa} benchmark to evaluate all the baselines and our proposed Ego Scene Augmentation under a consistent, unified, and carefully standardized experimental protocol. For more details about this benchmark, please refer to the \textit{Supplementary Material}.

\input{tables/ablation_components}

\subsection{Quantitative Results}

\noindent We present the evaluation results on the EgoTextVQA~\cite{zhou2025egotextvqa} benchmark in Table~\ref{tab:egotextvqa_result_indoor} and Table~\ref{tab:egotextvqa_result_outdoor}. As shown in Table~\ref{tab:egotextvqa_result_indoor} and Table~\ref{tab:egotextvqa_result_outdoor}, our proposed Ego Scene Augmentation reasoning framework demonstrates strong performance improvement, 
achieving an \textbf{8.14\%} gain in accuracy and a \textbf{4.27\%} gain in score on the indoor setting as well as an \textbf{8.72\%} gain in accuracy and a \textbf{4.5\%} gain in score on the outdoor setting.
These substantial gains highlight the effectiveness of ESA in enhancing the reasoning capabilities of MLLMs for Egocentric VQA tasks.
The quantitative findings indicate that ESA exhibits competitive performance in the egocentric perspective reasoning, especially the spatial reasoning ability, achieving a total \textbf{0.19} score gain over existing reasoning methods for the indoor section and outdoor section.
The results provide evidence showing the effectiveness of our ESA in reinforcing the egocentric perspective reasoning for MLLMs.
Furthermore, our Ego-element Graph shows the most impressive performance improvement on the shopping subset, achieving a \textbf{15.91\%} gain in accuracy and a \textbf{7.61\%} gain in score on the shopping subset. 
Shopping scenes typically contain densely distributed objects, overlapping scene texts, and complex spatial layouts that require precise spatial reasoning. 
Conventional MLLMs mainly rely on global visual embeddings and thus often fail to distinguish fine-grained positional relations.
In contrast, our Ego-element Graph explicitly models the spatial and semantic relations among detected scene elements (objects, texts, and surfaces) through graph-based reasoning. 
This structured representation enables the model to localize and correlate relevant entities more accurately, resulting in more robust performance in cluttered, text-rich environments and the largest relative gain on the shopping subset.







\subsection{Qualitative Results}

As illustrated in Figure~\ref{fig:experiment qualitative results}, we present a visual comparison between MLLMs (InstructBLIP~\cite{dai2023instructblip}, Emu3~\cite{wang2024emu3}, LLaVA-NeXT~\cite{li2024llava}, InternVL2~\cite{chen2024internvl}, Qwen2-VL~\cite{wang2024qwen2}, Qwen2.5-VL~\cite{bai2025qwen2}) and our proposed ESA framework. The visual results demonstrate a notable improvement in spatial reasoning achieved by ESA in the egocentric perspective reasoning. For instance, given the question {\fontfamily{qcr}\selectfont["Where did I place the super glue?"]} and the key frame from the video, only ESA accurately identifies the correct position of the super glue, as shown in Figure~\ref{fig:experiment qualitative results}. Furthermore, in response to the question {\fontfamily{qcr}\selectfont["Where is more aluminum foil available for cooking?"]} and the key frame from the video, all conventional MLLMs fail to answer the position of the aluminum foil correctly (though the Score of LLaVA-NeXT-Video is 4, it just guesses the places where the aluminum foil is). In contrast, ESA not only identifies the text {\fontfamily{qcr}\selectfont"Sanita"} required for the answer, but also locates the position of the aluminum foil, as also shown in Figure~\ref{fig:experiment qualitative results}. These visual comparisons show the substantial gains in egocentric scene reasoning provided by our ESA framework. Additional results are presented in the \textit{Supplementary Material} to further validate the effectiveness of ESA.

%% file: tables/ablation_num.tex
\begin{table*}[t!]
\centering
\Large
\caption{\textbf{Ablation on the Maximum Number of Nodes.} We present an ablation study on the hyperparameter controlling the maximum number of nodes using the EgoTextVQA benchmark, systematically varying its value to 2, 4, 6, 8, and 10. The results indicate that a setting of 4 for \textit{maximum number of nodes} shows the best overall performance. ``ACC" refers to Accuracy, and ``Num. Object" to the maximum number of objects we detect.}
\resizebox{\textwidth}{!}{
\begin{tabular}{l|cc|cc|cc|cc|cc|cc|cc} 
\toprule
\multicolumn{15}{c}{\Large\textbf{Ablation on the Maximum Number of Nodes on EgoTextVQA-Indoor}} \\ 
\midrule
\multirow{2}{*}{\textbf{Num. Object}} &
\multicolumn{2}{c|}{\textbf{Average}} &
\multicolumn{2}{c|}{\textbf{Hands-on}} &
\multicolumn{2}{c|}{\textbf{Kitchen}} &
\multicolumn{2}{c|}{\textbf{Shopping}} &
\multicolumn{2}{c|}{\textbf{Gameplay}} &
\multicolumn{2}{c|}{\textbf{Book-Related}} &
\multicolumn{2}{c}{\textbf{Others}} \\ 
\cmidrule(lr){2-15}
 & \textbf{ACC} & \textbf{Score} 
 & \textbf{ACC} & \textbf{Score} 
 & \textbf{ACC} & \textbf{Score} 
 & \textbf{ACC} & \textbf{Score}
 & \textbf{ACC} & \textbf{Score} 
 & \textbf{ACC} & \textbf{Score} 
 & \textbf{ACC} & \textbf{Score} \\ 
\midrule
Value = 2  & 32.91 & 2.30 & 33.29 & 2.32 & 28.06 & 2.11 & 27.65 & 2.09 & 43.85 & 2.73 & 36.22 & 2.43 & 27.36 & 2.08 \\
\rowcolor{gray!15}
Value = 4  & 33.74 & 2.34 & 33.86 & 2.35 & 28.96 & 2.15 & 28.49 & 2.12 & 45.00 & 2.78 & 36.84 & 2.45 & 28.86 & 2.14 \\
Value = 6  & 33.32 & 2.32 & 33.00 & 2.32 & 28.36 & 2.12 & 28.77 & 2.13 & 43.85 & 2.73 & 36.53 & 2.44 & 29.85 & 2.18 \\
Value = 8  & 33.41 & 2.32 & 33.14 & 2.32 & 28.66 & 2.14 & 28.49 & 2.12 & 44.19 & 2.74 & 36.53 & 2.44 & 29.85 & 2.18 \\
Value = 10 & 33.45 & 2.33 & 33.00 & 2.32 & 29.25 & 2.16 & 28.49 & 2.12 & 44.52 & 2.75 & 36.53 & 2.44 & 29.35 & 2.16 \\
\bottomrule
\end{tabular}
}
\label{tab:ablation_num}
\end{table*}

%% file: tables/ablation_components.tex
\begin{table*}[t!]
\centering
\Large
\caption{\textbf{Ablation on Components of the Pipeline.} We present an ablation study on components of the pipeline using the EgoTextVQA benchmark. ``ACC" refers to Accuracy. v4: all components.}
\resizebox{\textwidth}{!}{
\begin{tabular}{l|cc|cc|cc|cc|cc|cc|cc} 
\toprule
\multicolumn{15}{c}{\Large\textbf{Ablation on Components of the Pipeline on EgoTextVQA-Indoor}} \\ 
\midrule
\multirow{2}{*}{\textbf{Component}} &
\multicolumn{2}{c|}{\textbf{Average}} &
\multicolumn{2}{c|}{\textbf{Hands-on}} &
\multicolumn{2}{c|}{\textbf{Kitchen}} &
\multicolumn{2}{c|}{\textbf{Shopping}} &
\multicolumn{2}{c|}{\textbf{Gameplay}} &
\multicolumn{2}{c|}{\textbf{Book-Related}} &
\multicolumn{2}{c}{\textbf{Others}} \\ 
\cmidrule(lr){2-15}
 & \textbf{ACC} & \textbf{Score} 
 & \textbf{ACC} & \textbf{Score} 
 & \textbf{ACC} & \textbf{Score} 
 & \textbf{ACC} & \textbf{Score}
 & \textbf{ACC} & \textbf{Score} 
 & \textbf{ACC} & \textbf{Score} 
 & \textbf{ACC} & \textbf{Score} \\ 
\midrule
v0 Original              & 31.20 & 2.24 & 31.81 & 2.27 & 26.87 & 2.07 & 24.58 & 1.97 & 41.53 & 2.65 & 36.02 & 2.46 & 24.88 & 1.98 \\
v1 + Region           & 31.38 & 2.25 & 31.13 & 2.24 & 25.97 & 2.03 & 27.09 & 2.07 & 42.52 & 2.68 & 35.29 & 2.40 & 25.87 & 2.02 \\
v2 + Depth        & 33.41 & 2.32 & 33.57 & 2.33 & 27.46 & 2.09 & 29.05 & 2.14 & 44.52 & 2.76 & 36.53 & 2.44 & 28.86 & 2.14 \\
v3 + Text & 33.54 & 2.33 & 33.43 & 2.33 & 28.06 & 2.11 & 28.21 & 2.11 & 45.51 & 2.79 & 37.15 & 2.46 & 28.86 & 2.14 \\
\rowcolor{gray!15}
v4 + Organize Information in Graph            & 33.74 & 2.34 & 33.86 & 2.35 & 28.96 & 2.15 & 28.49 & 2.12 & 45.00 & 2.78 & 36.84 & 2.45 & 28.86 & 2.14 \\
\bottomrule
\end{tabular}
}
\label{tab:ablation_components}
\end{table*}

%% file: sec/5_ablation_study.tex
\section{Ablation Study}
\label{sec:ablation study}

\begin{figure*}[t]
  \centering
  \includegraphics[width=0.58\textwidth]{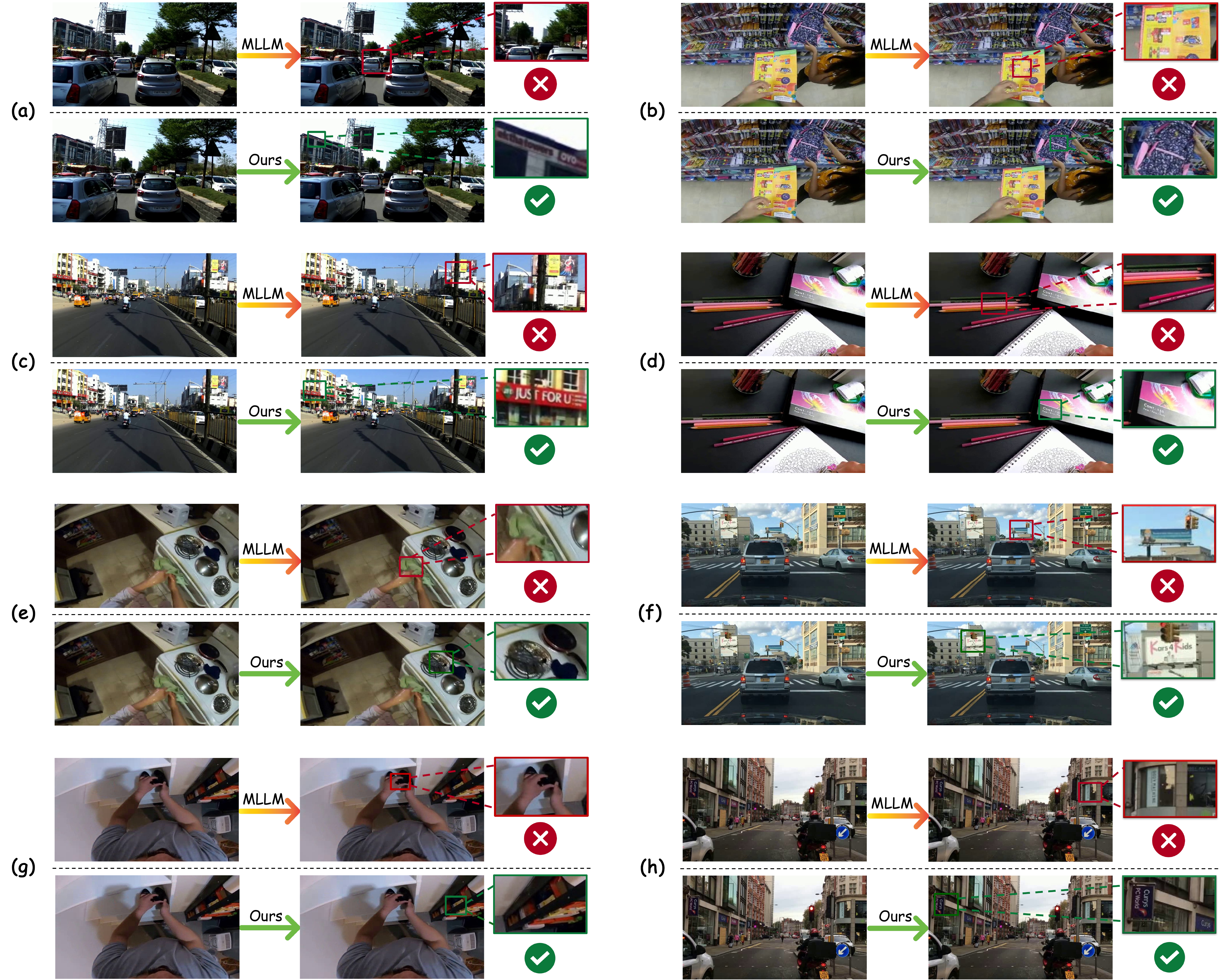}
  \vspace{-2pt}
  \caption{\textbf{Ablation on Components of the Pipeline.} Comparison of region-level recognition between existing MLLMs and our proposed Ego Scene Augmentation, showing that ESA produces more accurate and context-aware localization results.}
  \label{fig:ablation2}
\end{figure*}

\noindent \textbf{Ablation on the Maximum Number of Nodes.}
To determine the optimal value for the maximum-number-of-nodes hyperparameter, we experiment with settings of 2, 4, 6, 8, and 10. This hyperparameter determines how many detected nodes are retained in the Ego-element Graph, directly shaping the spatial context available for downstream reasoning. A smaller number may omit relevant cues that are critical for localizing targets or interpreting relations, while a larger number can introduce redundant or noisy nodes, making the Ego-element Graph unnecessarily cluttered and harder for the model to interpret.
The corresponding results are reported in Table~\ref{tab:ablation_num}. As indicated by the results, setting this hyperparameter to 4 yields the best performance on the EgoTextVQA~\cite{zhou2025egotextvqa} benchmark, suggesting that this configuration provides an effective balance between information sufficiency and graph sparsity, allowing the model to capture essential scene structure without being overwhelmed by additional irrelevant detections.

\noindent \textbf{Ablation on Components of the Pipeline.}
We conduct a comprehensive ablation to analyze the contribution of each module in our ESA pipeline, including \emph{Region}, \emph{Depth}, \emph{Text}, and \emph{Graph Organization}. Table~\ref{tab:ablation_components} summarizes quantitative results on EgoTextVQA-Indoor, while Figure~\ref{fig:ablation1} and Figure~\ref{fig:ablation2} provide qualitative comparisons.

Starting from the baseline MLLM (v0), progressively adding Region extraction (v1), Depth estimation (v2), Text recognition (v3), and finally organizing all cues into the Ego-element Graph (v4) leads to consistent and clearly observable performance gains. Average accuracy increases from \textbf{31.20} $\rightarrow$ \textbf{33.74}, and average score increases from \textbf{2.24} $\rightarrow$ \textbf{2.34}. Depth (v2) brings the largest single improvement, especially on \textit{Gameplay} and \textit{Others}. Text (v3) further enhances semantic grounding, particularly in \textit{Gameplay} and \textit{Book-Related}. Graph organization (v4) integrates multi-source cues into a unified spatial structure and yields consistent improvements across all categories, achieving the best results in \textit{Hands-on}, \textit{Kitchen}, and \textit{Others}.

\noindent Qualitative examples demonstrate the effect of each component in improving overall reasoning: \\
(1) In Figure~\ref{fig:ablation1}, combining the input with the Ego-element Graph allows the model to reason about 3D object relations and nearby text cues, enabling more consistently correct answers where the baseline MLLM fails. \\
(2) In Figure~\ref{fig:ablation2}, region-level comparisons show that ESA refines object localization and text grounding, correcting misrecognitions commonly observed in existing MLLMs.
Overall, the ablation indicates that depth cues, text information, and structured graph organization are highly complementary, collectively enhancing egocentric spatial reasoning in varied and complex scenes.

%% file: sec/6_conclusion.tex
\section{Conclusion}
\label{sec:conclusion}
In this paper, we introduce Ego Scene Augmentation (ESA), a plug-and-play egocentric spatial reasoning framework that strengthens the spatial perception of MLLMs without additional model training. ESA constructs an Ego-element Graph that explicitly organizes object semantics, visual regions, depth cues, and scene text, and serializes this structured evidence into an enhanced spatial prompt for inference. Experiments on the indoor and outdoor settings of EgoTextVQA demonstrate consistent improvements, with relative accuracy gains of \textbf{8.14\%} and \textbf{8.72\%}, respectively. Ablation studies show that depth contributes the largest individual improvement, while text cues and graph organization provide complementary benefits. Overall, explicit spatial evidence effectively complements the implicit visual representations of existing MLLMs in cluttered and text-rich egocentric scenes.

\noindent \textbf{Limitations and Future Work.} ESA remains sensitive to viewpoint changes, and errors from external perception modules may propagate into the final answer. Without camera calibration, it provides only ordinal rather than metric depth. Future work will pursue uncertainty-aware graphs, metric geometry, temporal aggregation, and broader tests.

%% file: sec/X_suppl.tex


\clearpage
\setcounter{page}{1}
\maketitlesupplementary

\setcounter{section}{0}
\renewcommand{\thesection}{\Alph{section}}
\renewcommand{\thesubsection}{\thesection.\arabic{subsection}}

\section{More Details of Ego Scene Augmentation}
\label{Appendix:More Details of Ego Scene Augmentation}

\subsection{More Details of the Reasoning Prompt}
\label{Appendix:Reasoning_Prompt}

\begin{figure*}[!t]
    \centering
    \includegraphics[width=\textwidth]{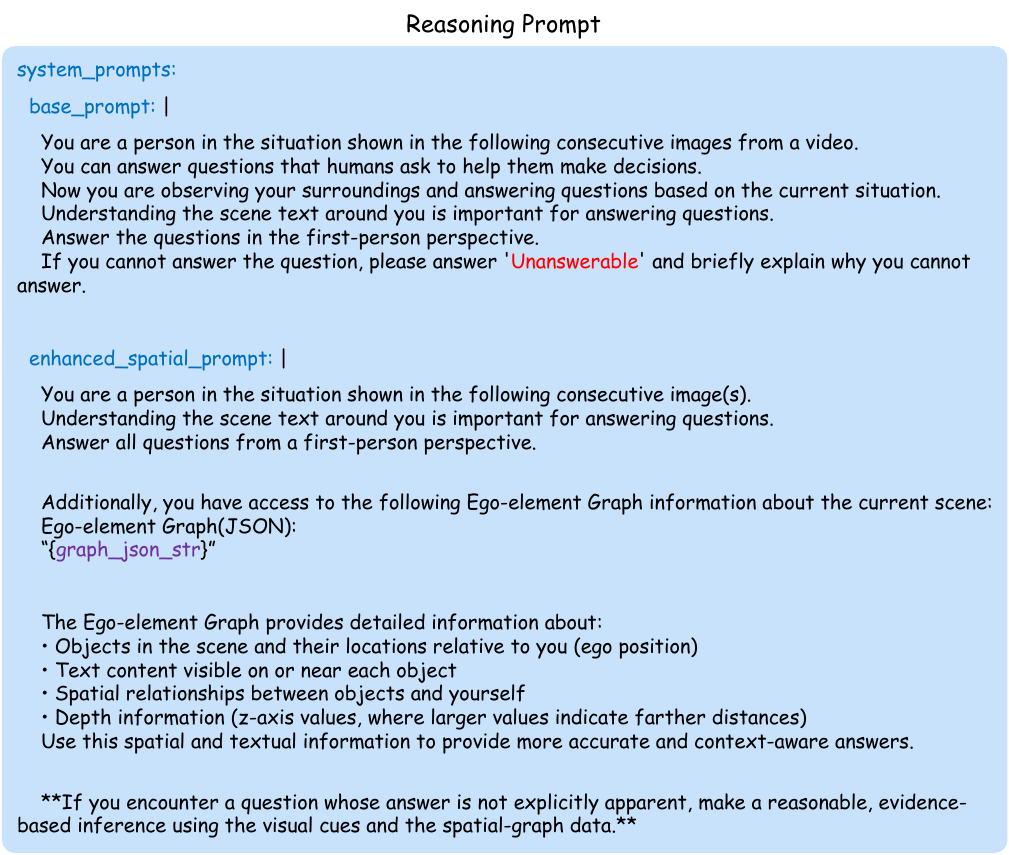}
    \caption{The full reasoning prompt}
    \label{fig:reasoning_prompt}
\end{figure*}

Figure~\ref{fig:reasoning_prompt} presents the full reasoning prompt used in our ESA framework.
While the base prompt guides the MLLM to answer from a first-person perspective, the reasoning
prompt is specifically designed to enhance the model’s spatial reasoning and contextual decision-making
during the answer-generation stage.
The reasoning prompt begins by placing the model in the role of the person who recorded the egocentric
video, emphasizing that the model must rely solely on the visual evidence contained in the provided
key frame. The prompt reminds the model to avoid hallucinated assumptions and to explicitly respond
``\texttt{Unanswerable}’’ when the image does not contain enough information. This encourages
evidence-grounded reasoning instead of speculative responses.
Furthermore, the reasoning prompt requires the model to carefully inspect the scene, identify objects of
interest, understand their spatial arrangement, and integrate scene text when necessary. When
combined with the enhanced spatial prompt derived from the Ego-element Graph, the reasoning prompt
ensures that the MLLM leverages structured spatial cues, depth information, and textual annotations to
form more accurate and context-aware answers.
Overall, this prompt plays a critical role in our ESA pipeline: it provides explicit and clear instructions for interpreting visual cues, grounding the reasoning process in egocentric context, and ensuring that the model uses both spatially organized information and available image evidence to produce reliable predictions.


\subsection{More Details of Key-Frame Selection}
\label{Appenix:More Details of Key-Frame Selection}

EgoTextVQA provides full egocentric video segments for each question, but the key visual
evidence often resides in a specific moment rather than across the entire clip. Following
this observation, we adopt a CLIP-based relevance scoring method to select a single
key frame per question. Given the question and all available frames prior to the annotated
timestamp, we compute question–frame similarity scores and select the frame with the
highest CLIP alignment. This key-frame formulation preserves the egocentric difficulty of
the benchmark while enabling consistent and computationally efficient reasoning within
our ESA pipeline.

\section{More Details of Experiment}
\label{Appendix:More Details of Experiment}

\subsection{More Details of Benchmark}
\label{Appendix:Details_of_benchmark}

\noindent \textbf{EgoTextVQA Benchmark}
is a large-scale egocentric scene-text aware VideoQA benchmark designed to evaluate
how well models understand dynamic first-person environments enriched with textual and
object-level cues. The dataset consists of approximately 1.5K egocentric videos and over 7K
question–answer pairs collected from two major domains: \emph{indoor household activities} and
\emph{outdoor driving or navigation}. Compared with static scene-text VQA datasets such as
TextVQA or ST-VQA, EgoTextVQA features significantly more challenging visual conditions,
including motion blur, partial visibility of text, cluttered backgrounds, and continuous viewpoint
changes inherent to head-mounted cameras.
A distinctive property of EgoTextVQA is its strong dependence on scene text. Around half of
the questions and answers explicitly require models to identify, read, or reason about text appearing
on objects, signs, or packaging materials within the scene. This makes the benchmark particularly
suited for evaluating methods that integrate text cues with object localization and spatial context.
The benchmark further annotates each question with a specific timestamp, ensuring that the question
is grounded in a realistic egocentric scenario where the camera wearer is actively performing a task.
For evaluation, EgoTextVQA categorizes questions into semantically meaningful types, including
Location, Direction, Description, Intention Reasoning, and others. These categories reveal diverse
aspects of egocentric perception—ranging from identifying object placement or reading labels to
understanding user intention or spatial relations within cluttered environments.
Although EgoTextVQA is provided as a video-based benchmark, many questions are highly
localized around a single informative moment in the video. Following common practice in recent
egocentric VQA systems, our framework adopts a \emph{key-frame formulation} where a relevant
frame is selected based on visual–textual alignment prior to applying our Ego-element Graph
construction. This preserves the dataset’s egocentric challenges—such as scene-text dependence and
complex spatial layout—while enabling consistent, frame-based reasoning within our ESA pipeline.
Overall, EgoTextVQA provides a rich and demanding evaluation environment for thoroughly assessing egocentric spatial reasoning. Its integration of real-world motion, diverse object layouts, and notable scene-text dependency makes it a strong and reliable testbed for evaluating the effectiveness of our proposed ESA framework.

\subsection{Evaluation Protocol}
\label{Appendix:Evaluation Protocol}

We strictly follow the official evaluation protocol of EgoTextVQA, reporting both Accuracy
and the semantic Score metric defined in the benchmark. All baselines and our ESA model
are evaluated using identical key-frame inputs, ensuring fair and consistent comparison
across methods.

\section{Additional Visual Results}

\begin{figure*}[!t]
    \centering
      
    \includegraphics[width=\textwidth]{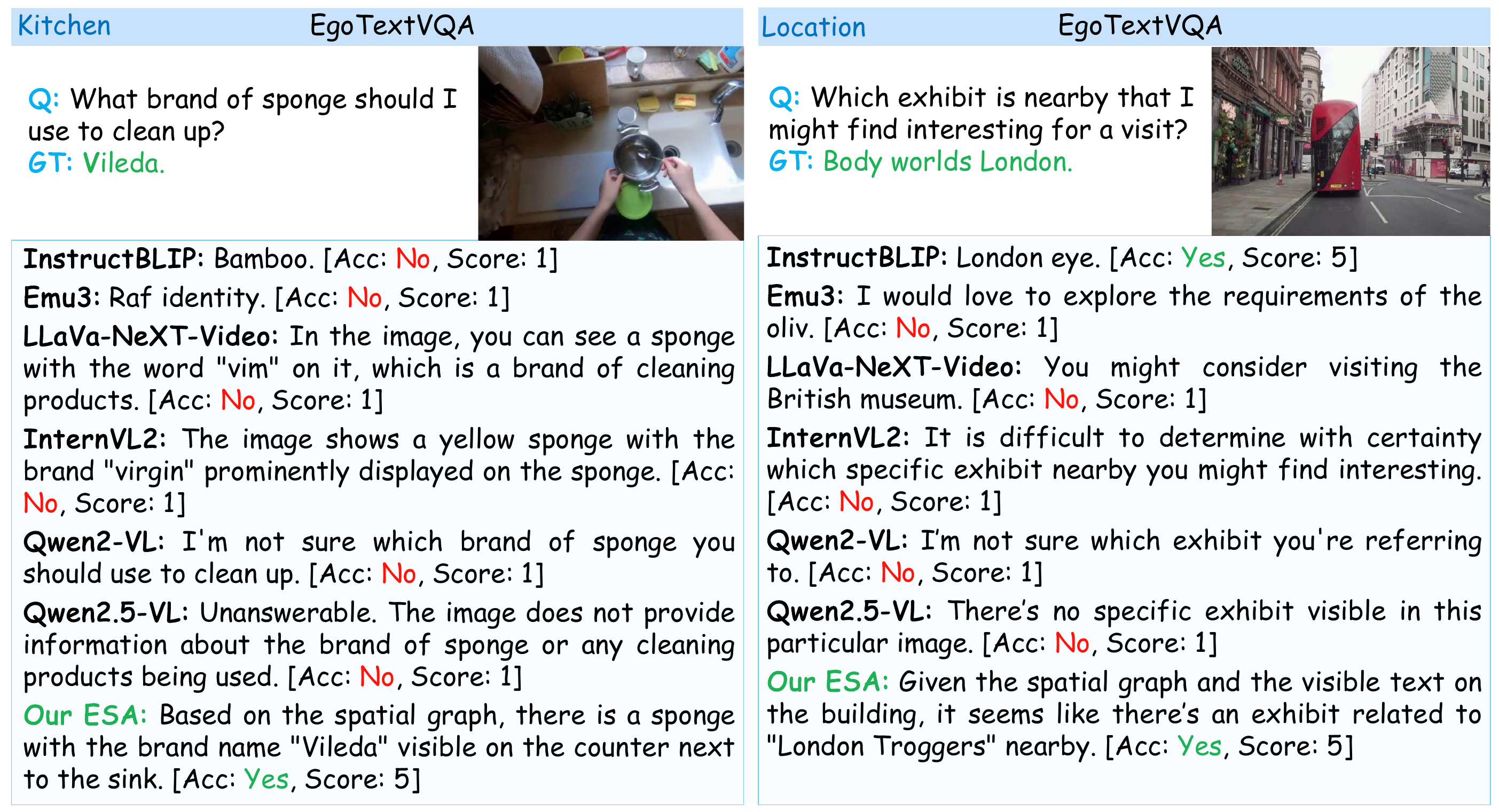}
    \includegraphics[width=\textwidth]{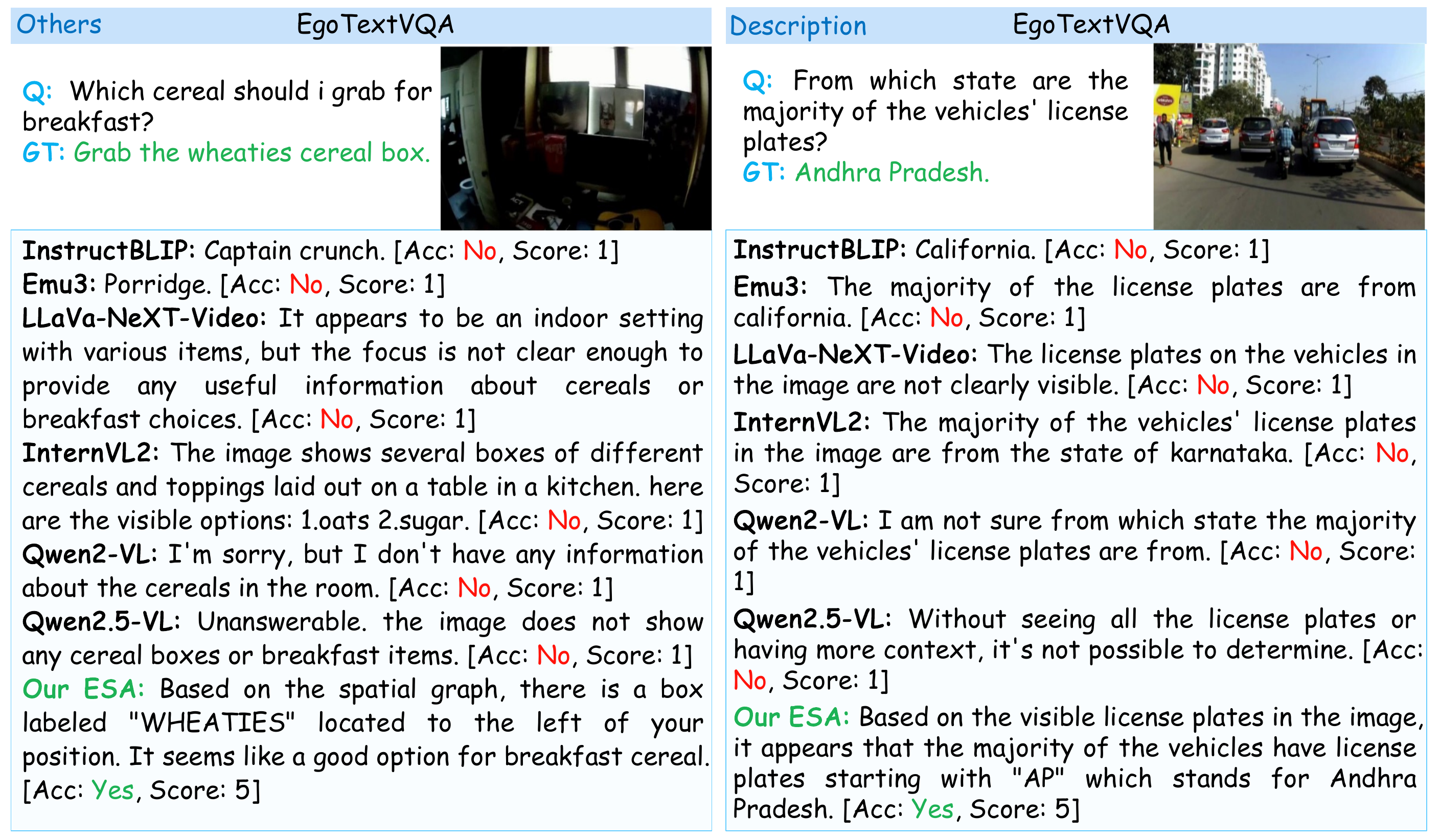}
    \caption{Additional visual comparison with ESA (group 1).}
    \label{fig:additional_visual_comparison_group1}
\end{figure*}

\begin{figure*}[t!]
  \centering

  \includegraphics[width=\textwidth]{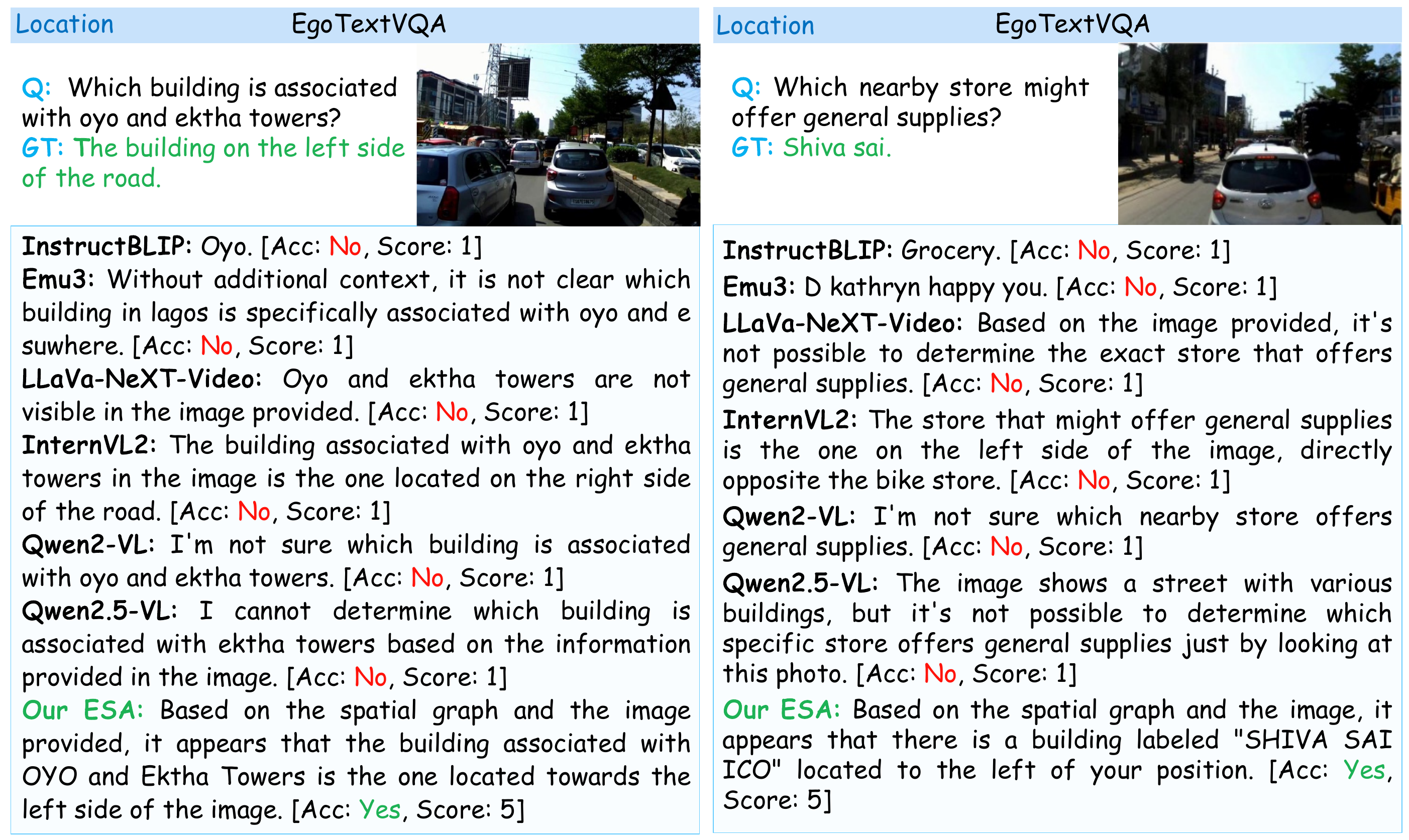}
  \includegraphics[width=\textwidth]{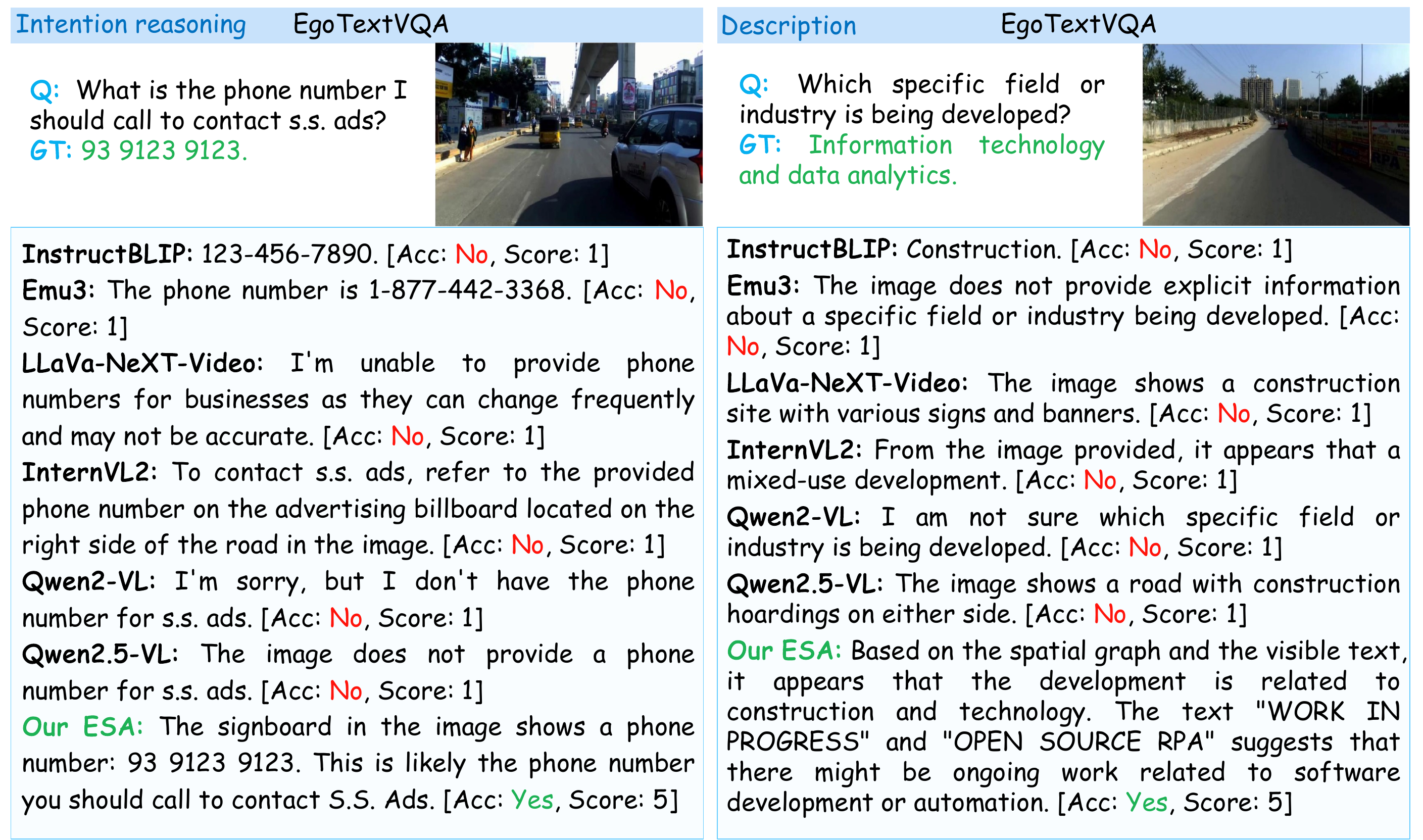}
  \caption{Additional visual comparison with ESA (group 2).}
  \label{fig:additional_visual_comparison_group2}
\end{figure*}

We present additional visual comparisons in Figures~\ref{fig:additional_visual_comparison_group1} and Figure~\ref{fig:additional_visual_comparison_group2},  which clearly illustrate the substantial performance improvements achieved by ESA over existing base MLLMs. Across diverse and challenging egocentric scenes, ESA enables the model to correctly localize queried objects and ground scene text, while the base MLLMs often fail to resolve spatial relations and instead output incorrect predictions. These qualitative examples further highlight how the explicit spatial cues provided by the Ego-element Graph lead to more accurate and well-grounded answers in challenging egocentric settings.

\section{The Use of Large Language Models(LLM)}
We used OpenAI's GPT-5 to assist with the careful refinement and proofreading of certain sentences in this paper. The LLM was used exclusively to enhance the clarity and overall coherence of our writing. All substantive content contributions are fully made by the authors.

\clearpage
\newpage